\documentclass{article}

\usepackage[preprint,nonatbib]{neurips_2026}
\usepackage[numbers,sort&compress]{natbib}

\usepackage[utf8]{inputenc}
\usepackage[T1]{fontenc}
\usepackage{hyperref}
\usepackage{url}
\usepackage{booktabs}
\usepackage{amsfonts}
\usepackage{amsmath}
\usepackage{amssymb}
\usepackage{amsthm}

\usepackage{nicefrac}
\usepackage{microtype}
\usepackage{xcolor}
\usepackage{graphicx}
\usepackage{subfigure}
\usepackage{algorithm}
\usepackage{algorithmic}
\usepackage{multirow}
\usepackage{pifont}
\usepackage{cleveref}
\usepackage{siunitx}
\usepackage{xcolor}
\usepackage{colortbl}
\newcommand{\B}[1]{\textbf{#1}}
\usepackage{tabularx}

\newcommand{\R}{\mathbb{R}}
\newcommand{\E}{\mathbb{E}}

\newcommand{\ba}{\mathbf{a}}
\newcommand{\bo}{\mathbf{o}}
\newcommand{\btheta}{\boldsymbol{\theta}}
\newcommand{\bdelta}{\boldsymbol{\delta}}
\DeclareMathOperator*{\argmin}{arg\,min}
\DeclareMathOperator*{\argmax}{arg\,max}

\newcommand{\Hu}[1]{\textcolor{red!70!black}{[\textbf{Hu:} #1]}}

\title{Test-time Adversarial Takeover: A Real-time Hijacking Interface against Robotic Diffusion Policies}

\author{%
    Zi Yin$^{*\ddagger}$ \\
    Tsinghua University
    \And
    Peilin Chai$^{*}$ \\
    Independent Researcher
    \And
    Siyuan Huang$^{*}$ \\
    Johns Hopkins University
    \And
    Zhanhao Hu$^{\dagger}$ \\
    UC Berkeley
}

\begin{document}

\maketitle

\renewcommand{\thefootnote}{\fnsymbol{footnote}}
\footnotetext[1]{Equal contribution}
\footnotetext[3]{Project lead}
\renewcommand{\thefootnote}{\arabic{footnote}}
\renewcommand{\thefootnote}{\fnsymbol{footnote}}
\footnotetext[2]{Correspondence: \texttt{huzhanhao@berkeley.edu}}
\renewcommand{\thefootnote}{\arabic{footnote}}

\begin{abstract}
Diffusion-based action generation has become a foundational component of embodied AI, but its reliance on visual conditioning leaves deployed visuomotor policies vulnerable to adversarial manipulation. Most prior attacks focus on disruption: they perturb the observation stream to reduce task success or induce erratic behavior. We study a stronger threat, Test-time Adversarial Takeover (TAKO), in which an attacker obtains a real-time steering interface over a frozen robot policy and turns it into a remotely piloted instrument. TAKO learns a small vocabulary of reusable universal patches through differentiable diffusion inference; at test time, the attacker switches among these patches in the camera stream to compose attacker-chosen trajectories. This works because the perturbation acts on the visual conditioning pathway, where the induced bias can persist through iterative generative inference. We further show that the natural targeted baseline, target-policy matching, fails because the victim policy cannot reliably supervise itself on out-of-distribution target shifts. Across four tasks (2D manipulation, simulated aerial delivery, simulated ground navigation, and physical-world ground navigation), two visual encoders (ResNet-18 and EfficientNet-B0 + Transformer), and three generative inference families (DDPM, DDIM, and flow matching), human operators achieve 100\% takeover success on attacker-defined objectives in every evaluated setting. The project page is available at \url{https://tako-attack.github.io}.

\end{abstract}

\section{Introduction}
\label{sec:intro}

Diffusion-based action generation has become a foundational building block in embodied AI. Since the introduction of Diffusion Policy~\citep{chi2023diffusion}, diffusion models have been widely adopted as critical action generation modules across a growing range of robotic systems, including vision-language-action (VLA) models and imitation learning policies.
Recently, vision has dominated many robotic policy learning tasks because it captures rich, fine-grained spatiotemporal information without explicit programming. In practice, virtually all deployed policies rely on image encoders such as ResNets, ViTs, or similar backbones to condition action generation on visual observations.
Consequently, vision inputs form a natural perturbation surface, so that any policy that conditions on visual input is inherently susceptible to adversarial manipulation through that channel. 

\begin{figure}[t]
  \centering
  \includegraphics[width=0.9\linewidth]{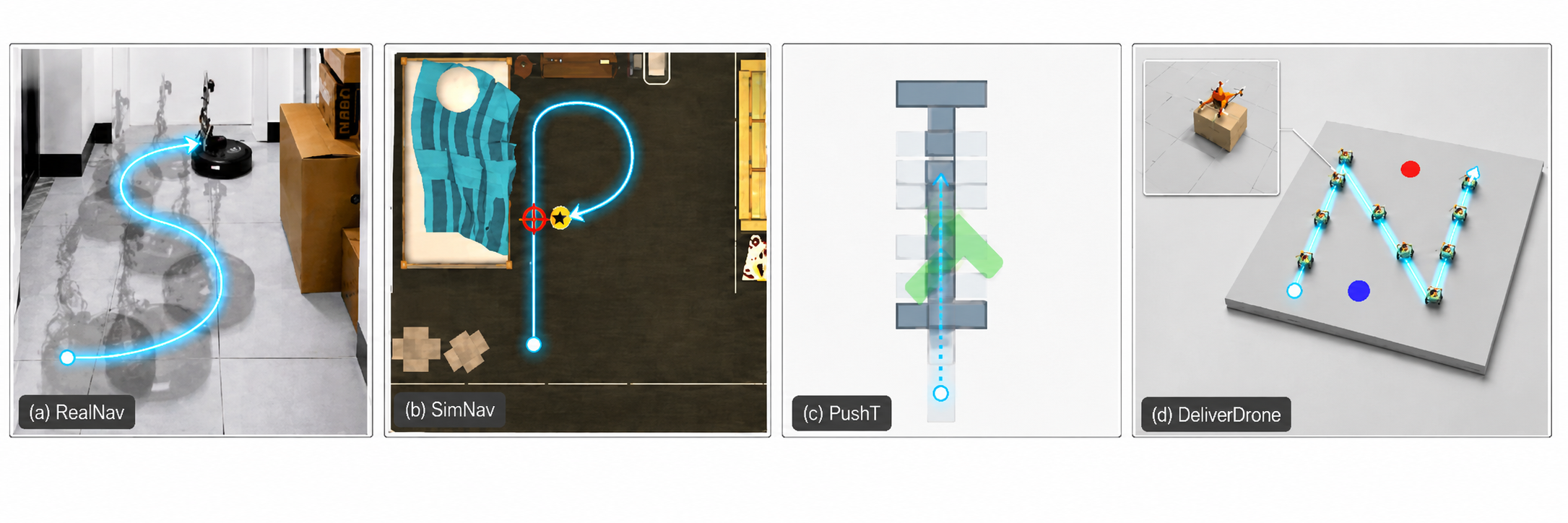}
  \setlength{\abovecaptionskip}{1pt}
  \caption{\textbf{Real-time adversarially steering imitation learning robots to follow arbitrary trajectories.} The tasks include: (a) RealNav, (b) SimNav, (c) PushT, (d) DeliverDrone. The trajectories are controlled by a human operator and resemble four letters: S, P, I, and N.}
  \label{fig:teaser}
\end{figure}

Existing adversarial attacks on robot policies have explored this vulnerability, but their threat models are limited to causing task failure or triggering task execution via language prompts in VLA systems. 
These attacks are undoubtedly important in safety-critical contexts, but they are quite limited in scope. Subverting a policy can cause failure, but it does not have the potential to cause targeted and sustained consequences. Language prompts can trigger unintended task execution, but they can only be applied to VLA models, cannot achieve out-of-distribution target trajectories, and can be detected by semantic checks on the language input.

A more severe threat is the potential to perform an arbitrary task that exceeds the existing learned policy. A straightforward and conventional attack formulation of this threat is to directly match the actions of a target-shifted policy that achieves the attacker's desired trajectory. We refer to this as the Target-Policy Matching (TPM) attack. However, it is hard under the TPM formulation: 1. The target policy is agnostic. The attacker must identify a target policy to steer the victim robot, which may be out-of-distribution for the victim robot. It is very likely for the victim robot to fail to generalize to the shifted target configuration. 2. Real-time constraint for test-time deployment. The attacker must steer the victim robot on a specific trajectory over a long time horizon. Since the attack must be effective at test time, the attacker typically cannot rely on iterative optimization or online feedback to refine it. However, relying on open-loop static attack intervention can lead to cumulative error and often fails to achieve the target goal.

We demonstrate a fundamentally different attack formulation, \textbf{T}est-time \textbf{A}dversarial Ta\textbf{k}e\textbf{O}ver (TAKO), to identify this class of threat. Rather than disrupting a deployed policy, an attacker can \emph{take it over}. By switching among a small set of pre-optimized universal adversarial patches injected into the camera stream at test time, the attacker obtains a real-time steering interface over the policy's output. 
The universal patches serve as a reusable action vocabulary: each patch corresponds to an action-primitive bias, such as ``move forward'' or ``move left'', and switching among them composes a trajectory that can reach arbitrary attacker-defined targets in real time.
The victim robot transforms from an autonomous agent into a remotely piloted instrument under the attacker's continuous control. The attack can be initiated at any time: once the adversarial patch is presented to the camera, the policy continues to generate smooth, executable action sequences that the attacker can control. This changes the deployment concern from accidental failure to unauthorized takeover — a critical shift from a safety problem to a security problem. As long as the attack is engaged, an independent operator can steer the robot to arbitrary attacker-defined targets in real time by switching among the universal patches, without the need to solve a complex optimization problem or generalization challenge at test time.
See \Cref{fig:teaser} for a demonstration of such an attack.

The vulnerability stems from the architecture of diffusion policies. The adversarial perturbation acts on the conditioning signal (visual features), so the iterative diffusion denoising process cannot correct the induced bias, but is rather steered by the adversarial input. Each patch is optimized with multiple states sampled in the environment, so the resulting bias becomes a universal and reusable action-primitive that is effective across different states.

We instantiate TAKO on diffusion-based imitation learning policies across three virtual tasks and one physical task (four tasks demonstrated in \Cref{fig:teaser}): a physical world ground navigation task (RealNav), a simulated ground navigation task (SimNav), a 2D manipulation task (PushT), and a simulated drone delivery task (DeliverDrone). In this setting, the robot's autonomous success at executing the attacker-defined target policy is nearly $0\%$, whereas human operators using TAKO achieve $100\%$ success in steering the robot to the same targets through our patch-vocabulary-based interface. The conventional TPM baseline fails entirely because the victim policy cannot generalize to shifted target configurations.


Our contributions are summarized as follows:
\begin{itemize}
    \item We introduce TAKO as a new adversarial paradigm for robot policy risks, shifting the objective from causing failure to obtaining sustained, operator-controllable authority over policy behavior (See \Cref{tab:paradigm} for the comparison with prior work).
    \item We show that diffusion-based visuomotor policies admit a simple steering interface: a pre-optimized vocabulary of adversarial patches that can be switched online to compose arbitrary attacker-directed trajectories.
    \item We demonstrate that this takeover paradigm is not task or architecture-specific, validating it across multiple tasks in virtual and physical environments, spanning distinct visual encoders (ResNet-18 and EfficientNet-B0+Transformer) and three generative inference families (DDPM, DDIM, flow matching).
\end{itemize}

\begin{table}[t]
  \caption{Comparison of the relevant attacks with TAKO}
  \label{tab:paradigm}
  \centering
  \small
  \setlength{\tabcolsep}{3.5pt}
  \resizebox{\columnwidth}{!}{%
  \begin{tabular}{lccccc}
    \toprule
    Demonstrated property
      & \shortstack{Enchanting{\scriptsize\citep{lin2017tactics}}}
      & \shortstack{DP-Attacker{\scriptsize\citep{chen2024dpattacker}}}
      & \shortstack{Jones et al.{\scriptsize\citep{jones2025adversarial}}}
      & \shortstack{Dirty-Road{\scriptsize\citep{sato2021dirty_road}}}
      & \textbf{Ours} \\
    \midrule
    Physical realizability & \ding{55} & \ding{51} & \ding{55} & \ding{51} & \ding{51} \\
    Diffusion-policy attack & \ding{55} & \ding{51} & \ding{55} & \ding{55} & \ding{51} \\
    Target-policy achieving & \ding{55} & \ding{55} & \ding{51} & \ding{55} & \ding{51} \\
    Real-time Intervention & \ding{55} & \ding{55} & \ding{55} & \ding{55} & \ding{51} \\
    Composable action vocabulary & \ding{55} & \ding{55} & \ding{55} & \ding{55} & \ding{51} \\
    \bottomrule
  \end{tabular}%
  }
\end{table}

\section{Related Work}
\label{sec:related}
\textbf{Diffusion models for robot action generation.}
Diffusion Policy~\citep{chi2023diffusion} established diffusion-based action generation as a core paradigm for visuomotor behavior cloning. Subsequent work expanded this family to new inputs, architectures, and faster inference variants (including flow matching)~\citep{ze2024dp3,ke2024diffuser_actor,prasad2024consistency,zhang2025flowpolicy,onedp2024,zhu2025scaledp,liu2025rdt1b}. Related generative action heads also appear in generalist robot policies and VLAs~\citep{ghosh2024octo,black2024pi0,black2025pi05,li2024cogact,hou2024dit_policy}. Because these systems all rely on visual conditioning to drive action generation, vulnerabilities in that pathway can transfer across embodiments and inference families.

\textbf{Adversarial patches in visual perception.}
Universal adversarial patches were introduced for image classifiers~\citep{brown2017adversarial} and later made physically robust via EOT-style optimization~\citep{athalye2018synthesizing,eykholt2018robust}. Follow-up work attacked detection and dense prediction pipelines~\citep{liu2019dpatch,thys2019fooling,xu2020adversarial_tshirt,yamanaka2020adversarial_depth,cheng2022physical_depth,ranjan2019attacking_flow,hu2022adversarial,hu2023physically} and modern encoders~\citep{fu2022patchfool,zhou2023advclip}. In robotics control, prior patch attacks mostly target low-dimensional steering outputs or lane-departure-style objectives~\citep{sato2021dirty_road,kong2020physgan,pavlitskaya2020feasibility}.

\textbf{Adversarial attacks on robot policies.}
Prior adversarial robotics work focuses on disruption: reducing task success or inducing failures under perturbed observations. Early results were shown in deep RL~\citep{huang2017adversarial,lin2017tactics,zhang2020robust}, and later extended to imitation and diffusion policies~\citep{chen2024dpattacker,patil2025vulnerable}. In VLA systems, both patch attacks and jailbreak-style prompting can alter behavior~\citep{wang2025exploring,jones2025adversarial,zhang2025badrobot,robey2024robopair}, while backdoor approaches require training-time compromise~\citep{kiourti2020trojdrl,wang2024trojanrobot}.

\section{Method}
\label{sec:method}
 

\subsection{Problem Formulation and Threat Model}
\label{sec:threat}

We consider a diffusion-based visuomotor policy $\pi_{\btheta}$ that maps an observation $\bo_t = (I_t, s_t)$, comprising an RGB image $I_t \in \R^{H_I \times W_I \times 3}$ and an optional proprioceptive state $s_t$, to an action sequence $\ba_{t:t+H_a} \in \R^{H_a \times d_a}$ over an execution horizon of $H_a$ steps. The action sequence is generated by a conditional diffusion process. Writing $\phi$ for the vision encoder and $\epsilon$ for the conditional denoiser (both subnetworks of $\pi_{\btheta}$), inference proceeds along a fixed chain
\begin{equation}
    I_t \;\xrightarrow{\;\phi\;}\; z_t \oplus s_t \;\xrightarrow\; c_t \;\xrightarrow{\;\epsilon,\, \tau_{1:K}\;}\; \hat{\ba}_{t:t+H_a},
    \label{eq:inference-chain}
\end{equation}
where $z_t = \phi(I_t)$ are the visual features, $c_t$ is the conditioning signal formed from $z_t$ concatenated with the proprioceptive state $s_t$ when present, and $\tau_{1:K}$ denotes a $K$-step deterministic sampler (DDPM, DDIM, or a forward-Euler integrator of a flow-matching velocity field). The policy parameters $\btheta$ are frozen after deployment, and the policy is operated in receding-horizon fashion: the first $H_e \leq H_a$ actions of $\hat{\ba}_{t:t+H_a}$ are executed, then the policy is re-invoked on the next observation.

The attacker has white-box access to $\pi_{\btheta}$ for offline patch optimization, and the ability to overlay a fixed-shape perturbation $\bdelta \in \R^{h \times w \times 3}$ at a fixed image location on each frame of the input stream at inference time. We write $\mathrm{apply}(I_t, \bdelta)$ for the patched image and, by extension, $\mathrm{apply}(\bo_t, \bdelta) = (\mathrm{apply}(I_t, \bdelta), s_t)$ for the patched observation. The perturbation is bounded under $\|\bdelta\|_\infty \leq \varepsilon$. The proprioceptive state $s_t$ is not perturbed; the model weights, training data, and task specification are not modified. Once the attack is engaged, the patch is composited into every observation that reaches the policy; once disengaged, the policy receives the clean stream again and resumes its trained behavior.

The conventional disruption threat model treats the attacker as an adversary who seeks a single perturbation $\bdelta$ that drives the policy off-task. Concretely, given a task-failure functional $\mathcal{L}_{\text{fail}}$ such as a negative task reward or distance to a forbidden region, the disruption attacker solves
\begin{equation}
    \bdelta^*_{\text{disrupt}} \;=\; \argmax_{\|\bdelta\|_\infty \leq \varepsilon} \; \E_{\bo \sim \mathcal{D}_{\text{deploy}}} \big[\, \mathcal{L}_{\text{fail}}\!\big(\pi_{\btheta}(\mathrm{apply}(\bo, \bdelta))\big) \,\big].
    \label{eq:disrupt-obj}
\end{equation}
The output of \Cref{eq:disrupt-obj} is a single perturbation that provides only an open-loop departure from the trained policy.

However, a more severe threat model allows the attacker to perform an arbitrary task by designing perturbations that guide the policy towards a desired outcome. It is much more difficult than disruption, because the attacker must solve for a perturbation that not only causes failure but also induces specific, targeted behavior. 

\begin{figure}[t]
  \centering
  \includegraphics[width=\linewidth]{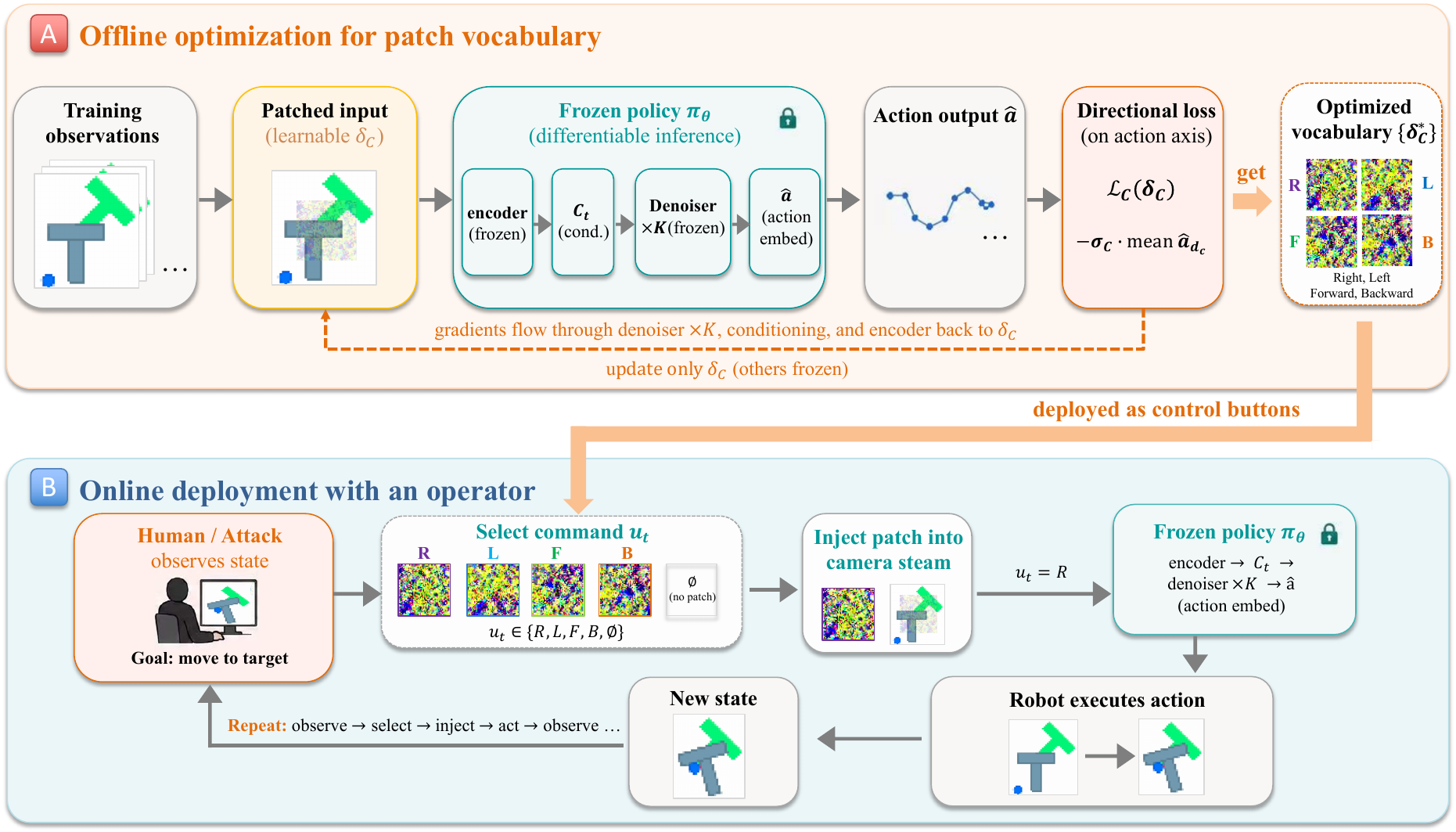}
  \caption{Attack pipeline. \textbf{Top:} offline universal-patch optimization via differentiable inference. Gradients flow from the directional loss through the frozen denoising chain and vision encoder back to the patch. \textbf{Bottom:} closed-loop online interactive deployment. The attacker selects a vocabulary entry via keyboard; the victim policy thus can be steered in real time.}
  \label{fig:pipeline}
\end{figure}

\subsection{A Natural Baseline: Target Policy Matching}
\label{sec:why-target-fails}

A straightforward approach to achieve the target policy is to directly optimize the perturbation such that the resulting action matches the desired reference policy. We refer to this approach as \emph{target policy matching} (TPM).
Concretely, let $\bo^{\text{shift}_c}$ denote an observation in which the goal has been displaced along direction $c$, and let
\begin{equation}
    \bar{\ba}^{\,c} \;=\; \pi_{\btheta}\!\big(\bo^{\text{shift}_c}\big)
    \label{eq:target-action}
\end{equation}
be the corresponding reference policy generated by querying the frozen victim policy model. The natural baseline TPM solves
\begin{equation}
    \bdelta_c^{\,\text{tgt}} \;=\; \argmin_{\|\bdelta_c\|_\infty \leq \varepsilon} \; \E_{\bo \sim \mathcal{D}} \,\Big\| \pi_{\btheta}\!\big(\mathrm{apply}(\bo, \bdelta_c)\big) \;-\; \bar{\ba}^{\,c} \Big\|_2^2.
    \label{eq:tgt-baseline}
\end{equation}
This is the standard targeted-attack template for the targeted threat model. 

However, this approach can easily fail in practice for two reasons. First, the reference policy $\bar{\ba}^{\,c}$ is generated by querying the victim policy, which may not produce the desired behavior when the goal is shifted, especially if the shifted goal is out-of-distribution. Second, even if the reference policy is generated successfully, the optimization in \Cref{eq:tgt-baseline} may fail to find a perturbation that reliably induces the desired behavior across the long horizon of the action sequence, especially when the policy is operated in receding-horizon mode and the accumulated drift from the reference trajectory causes the resulting policy runs to diverge from the target behavior. Given the context of real-time deployment, an online adaptation of the reference trajectory is not feasible. In practice, we find that TPM fails to produce a usable patch for any of our testbed of four tasks.

\subsection{TAKO Stage 1: Offline Optimization for Universal Patch Vocabulary}
\label{sec:optimization}

See \Cref{fig:pipeline} for the pipeline of TAKO. TAKO replaces reference matching with a control vocabulary: each command is a universal patch that induces a consistent action effect in the policy's output. Stage 1 learns this vocabulary offline so that each patch becomes a reusable control primitive. The patches are optimized by
\begin{equation}
    \bdelta_c^* \;=\; \argmin_{\|\bdelta_c\|_\infty \leq \varepsilon} \; \mathcal{L}_c(\bdelta_c).
    \label{eq:direction-loss}
\end{equation}
In general, $\mathcal{L}_c$ can be any loss that encourages the patch to induce a bias in the action output. We use a directional loss tailored to the tasks in our testbed, which require moving in four cardinal directions on a plane. For each command $c$, we fix an axis assignment $d_c \in \{1, \dots, d_a\}$ and a sign $\sigma_c \in \{+1, -1\}$ for each command $c$. Let $\hat{\ba}(\bo, \bdelta) = \pi_{\btheta}(\mathrm{apply}(\bo, \bdelta))$ denote the patched action sequence and let $\hat{a}_{t', d}(\bo, \bdelta)$ denote its $(t', d)$ entry. We define
\begin{equation}
    \mathcal{L}_c(\bdelta_c) \;=\; -\,\sigma_c \cdot \E_{\bo \sim \mathcal{D}} \, \frac{1}{H_a} \sum_{t'=1}^{H_a} \hat{a}_{t', d_c}(\bo, \bdelta_c),
    \label{eq:direction-loss}
\end{equation}
where $\mathcal{D}$ is an observation pool sampled from the training distribution, and the expectation is estimated by mini-batches in practice. For the four-direction vocabulary, $\mathcal{C} = \{\text{forward}, \text{backward}, \text{left}, \text{right}\}$ corresponds to two axes $d_c \in \{x, y\}$ and the two signs $\sigma_c \in \{+1, -1\}$.

Three properties make \Cref{eq:direction-loss} a clean replacement for \Cref{eq:tgt-baseline}. First, the loss is defined only on the single action-chunk output, so it never invokes the long-horizon policy on a shifted reference scene. Second, the action bias does not require a specific target, but rather only a consistent directional push, so the patch is not penalized for producing incoherent action sequences as long as the mean bias is in the right direction. Third, averaging over $\bo \sim \mathcal{D}$ makes the patch universal rather than trajectory-specific. 

To optimize \Cref{eq:direction-loss}, gradients are backpropagated through the full inference chain in \Cref{eq:inference-chain} to the patch. In practice, we unroll the sampler used at deployment time: DDIM~\citep{song2020denoising} for DDPM-trained policies and forward-Euler integration for flow-matching policies~\citep{lipman2023flow}. Because the patch acts on the visual conditioning pathway, the same recipe applies across score-based and flow-matching inference families.

\subsection{TAKO Stage 2: Online Compositional Deployment}
\label{sec:deployment}
 
Stage 2 turns that offline vocabulary into a real-time control interface. At deployment time, the attacker holds the optimized patches $\{\bdelta_c^*\}_{c \in \mathcal{C}}$ and selects an element $u_t \in \mathcal{C} \cup \{\varnothing\}$ at each replanning step. The patched observation
\begin{equation}
    \tilde{\bo}_t \;=\; \begin{cases} \mathrm{apply}(\bo_t, \bdelta_{u_t}^*) & u_t \in \mathcal{C} \\ \bo_t & u_t = \varnothing \end{cases}
    \label{eq:deploy-rule}
\end{equation}
is fed to the victim policy, which returns $\hat{\ba}_{t:t+H_a} = \pi_{\btheta}(\tilde{\bo}_t)$ and executes the first $H_e$ actions before the next replanning step. The attacker can choose $u_t$ from the realized state, but the policy itself only sees $\tilde{\bo}_t$ and treats it as clean input. When $u_t = \varnothing$, the policy immediately returns to normal behavior on the next chunk.
 
This protocol has two properties that distinguish takeover from conventional attack formulation. 
\emph{Composability}: the realized trajectory is determined by the command sequence $(u_1, u_2, \dots)$, so patches function as a switchable vocabulary rather than one-shot perturbations. \emph{Closed-loop control}: the attacker chooses $u_t$ online from the current state rather than committing to a fixed sequence of commands in advance, so the attack can correct for deviations from any single patch along the trajectory.

As long as the vocabulary is sufficiently expressive to reach the attacker's desired targets, the composability and closed-loop control of this protocol guarantee that the attacker can reach those targets regardless of how the policy drifts under each individual patch. The attacker's authority to steer online from the current state allows them to correct for any drift or misgeneralization of individual patches, so errors do not accumulate along the trajectory as they do under TPM.

\section{Experiments}
\label{sec:experiments}

\begin{figure}[t]
  \centering
  \includegraphics[width=0.8\linewidth]{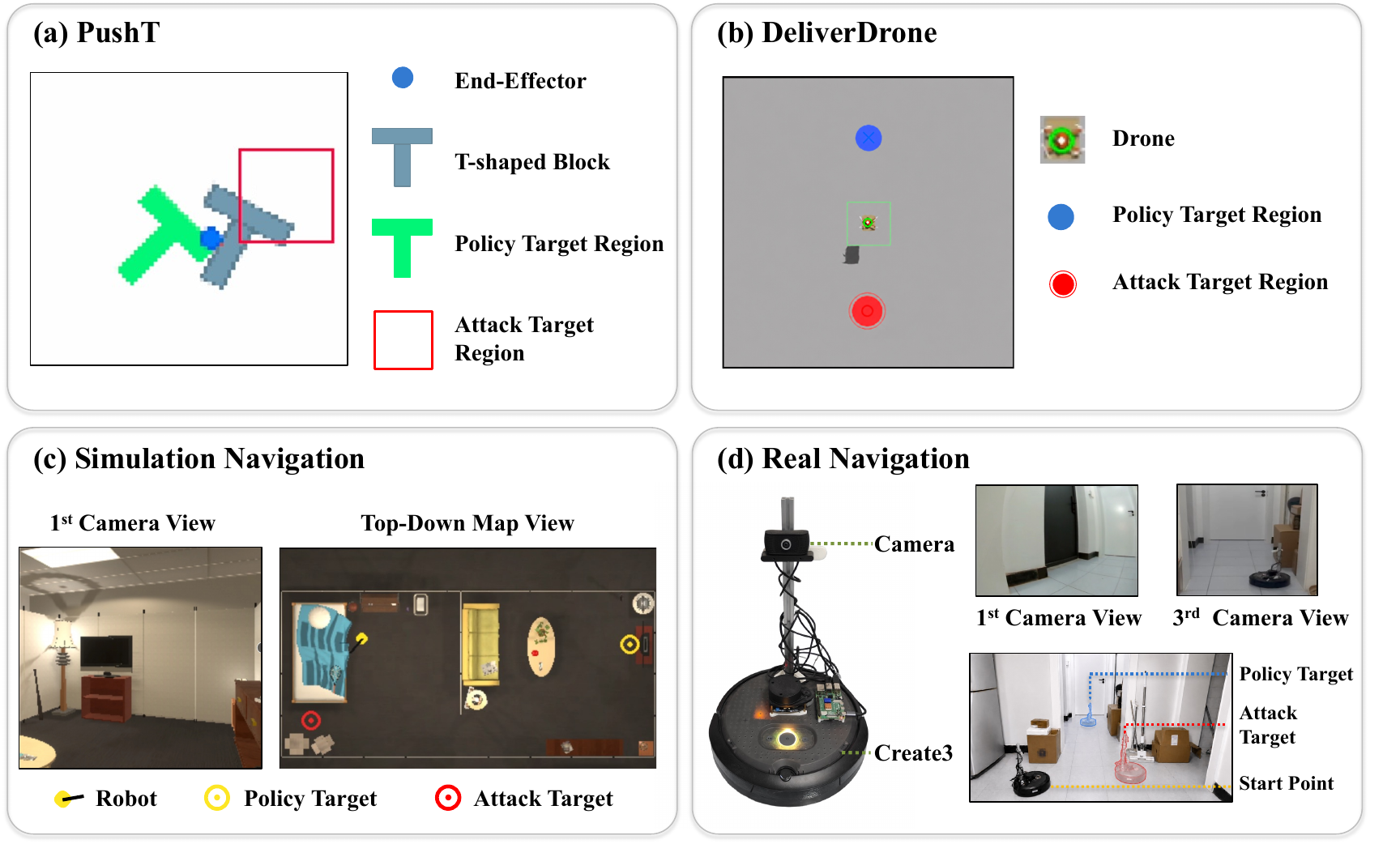}
  \setlength{\abovecaptionskip}{1pt}
  \caption{Illustration of the valuation tasks: (a)~PushT: 2D manipulation with a T-shaped block. (b)~DeliverDrone: simulated aerial delivery. (c)~SimNav: AI2-THOR indoor navigation with first-person and top-down views. (d)~RealNav: physical indoor navigation with an iRobot Create~3. The attack targets are only visible to human operators.}
  \label{fig:envs}
\end{figure}

\begin{table}[t]
  \caption{Setup overview for the tasks}
  \label{tab:setup}
  \centering
  \footnotesize
  \resizebox{\linewidth}{!}{%
  \begin{tabular}{lcccc}
    \toprule
     & PushT & DeliverDrone & SimNav & RealNav \\
    \midrule
    Environment & Sim. & Sim.\ (Gazebo) & Sim.\ (AI2-THOR) & Physical \\
    Embodiment & 2D agent & Sim.\ quadrotor & FPV agent & iRobot Create 3 \\
    Observation & 2$\times$96$^2$ RGB + pos & 4$\times$96$^2$ RGB + vel + goal & 2$\times$256$^2$ FPV + pos & 2$\times$96$^2$ FPV + pos \\
    Action space & 2D pixel velocity & 8 XY waypoints (m) & $(v, \omega)$ velocity & 2D body-frame waypoint \\
    Vision encoder & ResNet-18 & EffNet-B0 + Trans. & ResNet-18 & ResNet-18 \\
    Inference families tested & DDPM, DDIM, FM & DDPM, DDIM, FM & DDPM, DDIM, FM & DDPM, DDIM, FM \\
    Patch size & 48$\times$48 & 48$\times$48 & 128$\times$128 & 48$\times$48 \\
    Patch area ratio & 25\% & 25\% & 25\% & 25\% \\
    $\varepsilon$ & 64/255 & 64/255 & 64/255 & 64/255 \\
    \bottomrule
  \end{tabular}
  }
\end{table}

\subsection{Experimental Setup}
\label{sec:setup}

\textbf{Evaluation tasks.}
We evaluate TAKO on four diverse tasks: PushT (2D manipulation), DeliverDrone (simulated aerial delivery), SimNav (simulated egocentric indoor navigation), and RealNav (physical-world ground navigation). The four tasks differ on five axes:
\textbf{(1)~Vision encoder}: ResNet-18 (PushT, SimNav, RealNav) versus EfficientNet-B0 + 4-layer Transformer (DeliverDrone).
\textbf{(2)~Generative inference family}: every task is evaluated under DDPM, DDIM, and flow matching.
\textbf{(3)~Embodiment}: 2D manipulation agent, simulated quadrotor, first-person-view navigation agent, and a real navigation robot( iRobot Create~3 chassis).
\textbf{(4)~Action parameterization}: 2D pixel velocity (PushT, position-style), metric XY waypoints (DeliverDrone, position-style), continuous velocity $(v, \omega)$ (SimNav, velocity-style), and body-frame waypoints (RealNav, position-style). The inclusion of a velocity-action task is deliberate---it tests whether the attack transfers across position-style and velocity-style action spaces, which respond to a fixed action perturbation in qualitatively different ways.
\textbf{(5)~Environment}: simulation (PushT, DeliverDrone, SimNav) versus the physical world (RealNav).
\Cref{fig:envs} shows the task environments and \Cref{tab:setup} summarizes the configuration; per-task implementation details are in \Cref{sec:app_task}.

\textbf{Human interactive takeover protocol.}
See \Cref{fig:realnav,fig:drone} for the human task interface. Four researchers act as adversarial operators. Each trial presents an environment with the policy's original target and an attacker-defined target visible only to the operator (e.g., a blue marker placed at a predetermined position different from the original target). The operator steers the robot toward the target attacker in real time by pressing keyboard keys bound to pre-optimized universal patches. The patch is applied to the robot's visual input only when the corresponding key is held. We report success rate across all operators and trials (10 trials per operator per task, 40 trials total per task).

\textbf{Additional experimental setup} See \Cref{sec:add_exp_setup} for 
Policy implementation details, Patch vocabulary optimization details, and task implementation details.
\subsection{Patch Vocabulary Effectiveness}
\label{sec:patch_effectiveness}

\begin{table}[t]
  \caption{Patch directional success rate/\% (mean over 4 directions, with 95\% Wilson confidence intervals)}
  \label{tab:patch}
  \centering
  \small
  \begin{tabular}{lcccc}
    \toprule
     & PushT & DeliverDrone & SimNav & RealNav \\
    \midrule
    DDPM (100 steps)      & 98.5 \,[96.6,\,99.5] & 99.5 \,[97.6,\,99.8] & 65.8 \,[60.1,\,70.8] & 75.5 \,[70.2,\,79.9] \\
    DDIM (3--8 steps)     & 99.0 \,[97.1,\,99.7] & 99.5 \,[97.6,\,99.8] & 69.3 \,[63.9,\,74.3] & 74.5 \,[69.5,\,79.3] \\
    Flow matching         & 99.2 \,[97.6,\,99.8] & 100.0 \,[98.7,\,100] & 70.3 \,[64.9,\,75.2] & 97.3 \,[94.8,\,98.6] \\
    \bottomrule
  \end{tabular}
\end{table}
We first quantify whether the optimized patches induce a consistent directional bias in the victim policy's action output, and whether this property transfers across architectures and inference procedures. The result certifies that all patches behave as a directional action vocabulary; the closed-loop adversarial behavior they support is then evaluated in \Cref{sec:main_results}.

For each (task, direction, inference family) tuple, we applied the corresponding patch to a held-out observation pool ($N = 300$ samples per task). For each sample, we performed policy inference, extracted the predicted action sequence, and computed the mean shift along the action axis aligned with the commanded direction (relative to the clean policy's output for the same observation). We report \emph{Directional success} as the fraction of samples whose mean shift averaged over the predicted action window matches the commanded direction.
Per-direction breakdowns are reported in \Cref{app:additional}.

\Cref{tab:patch} reports directional success across the four tasks and three inference families; each cell is averaged over the four cardinal directions. The patches induce above-chance directional bias in every cell.
This shows that the patch optimization algorithm robustly yields action-primitive vocabulary across different vision encoders and generative inference families.

\subsection{Interactive Takeover Results}
\label{sec:main_results}

We then evaluated whether a human operator, given the directional vocabulary as a real-time control interface, can drive the robot to attacker-defined targets that no single patch was optimized to reach. For each task, we applied a three-layer protocol:
(1) Verify the victim policy is competent on its original task;
(2) Deploy the TPM baseline and report its Attack Success Rate (ASR);
(3) Deploy TAKO in the human-operator protocol and report the ASR.
ASR is defined as the percentage of trials in which the agent successfully reaches the attacker-defined target.

\begin{table}[t]
  \caption{Task-level success rate}
  \label{tab:main_results}
  \centering
  \small
  \begin{tabular}{lcccc}
    \toprule
     & PushT & DeliverDrone & SimNav & RealNav \\
    \midrule
    Clean task performance              & 10/10$^\dagger$ & 10/10 & 20/20 & 20/20 \\
    \midrule
    TPM ASR           & 0/40     & 0/40    & 0/40    & 0/40  \\
    \textbf{TAKO ASR (human)}  & \textbf{40/40} & \textbf{40/40} & \textbf{40/40} & \textbf{40/40} \\
    \bottomrule
  \end{tabular}
    \\{\footnotesize $^\dagger$ Success defined as IoU score $\geq 0.85$.}
\end{table}

\Cref{tab:main_results} shows the results. It carries three observations:
First, every victim policy was competent on its nominal task (top row).
Second, the TPM baseline achieved an ASR of \emph{zero} across all four tasks.
Third, our action vocabulary, composed at deployment time by a human operator, achieved 40/40 ASR across all tasks; the robots reached the attacker's targets that had never been trained to reach.
We further analyze the underlying failure modes of TPM and compare it with TAKO in \Cref{sec:baseline_analysis}.

\subsection{Analysis of TPM and TAKO}
\label{sec:baseline_analysis}



The failure of TPM in fact provides experimental evidence of the two problems argued in \Cref{sec:why-target-fails}: (i) OOD-supervision pathology --- the reference policy obtained from the frozen policy on a goal-shifted scene does not encode the desired direction; and (ii) accumulative drift---a static reference, fit at training time, becomes stale as the robot evolves under closed-loop deployment. We investigate each problem in the PushT task separately.

\begin{figure}[t]
  \centering
  \includegraphics[width=\linewidth]{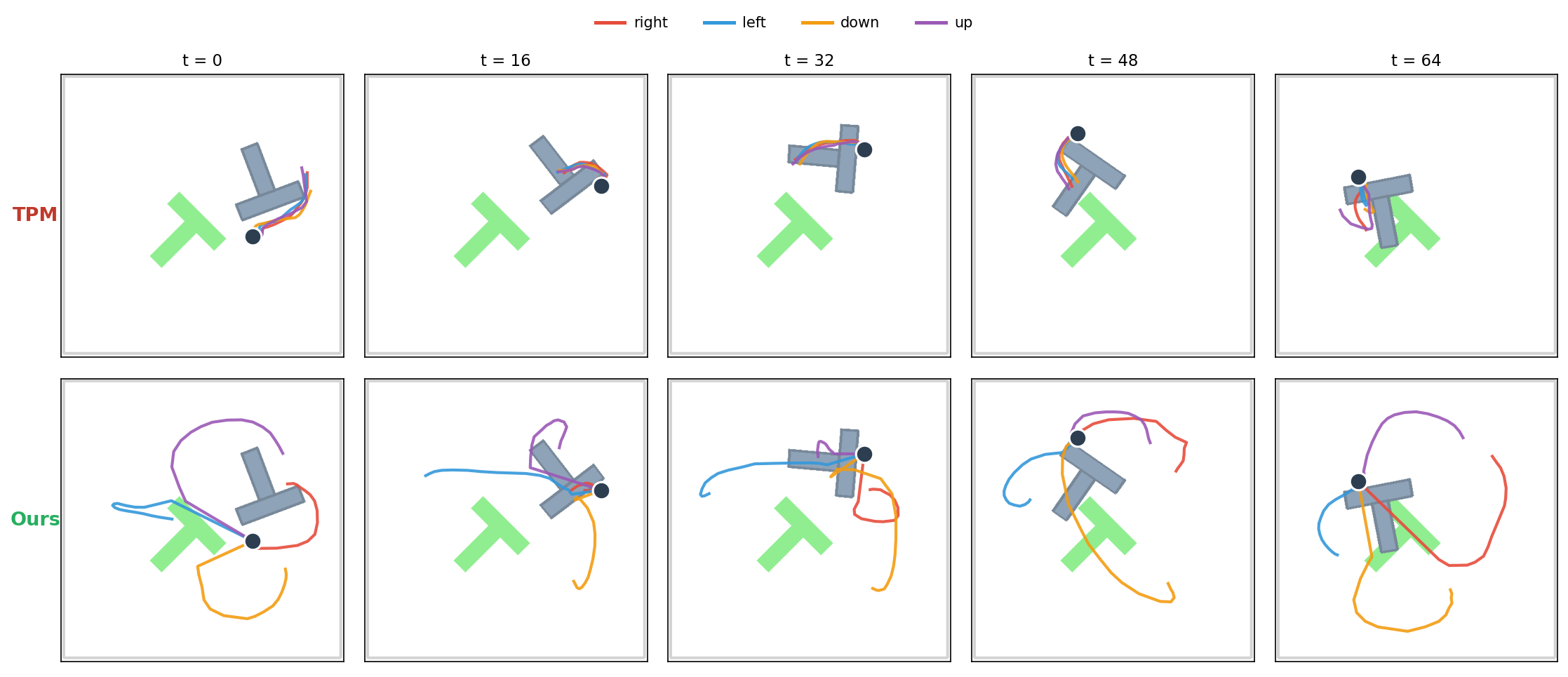}\\[5pt]
  \includegraphics[width=\linewidth]{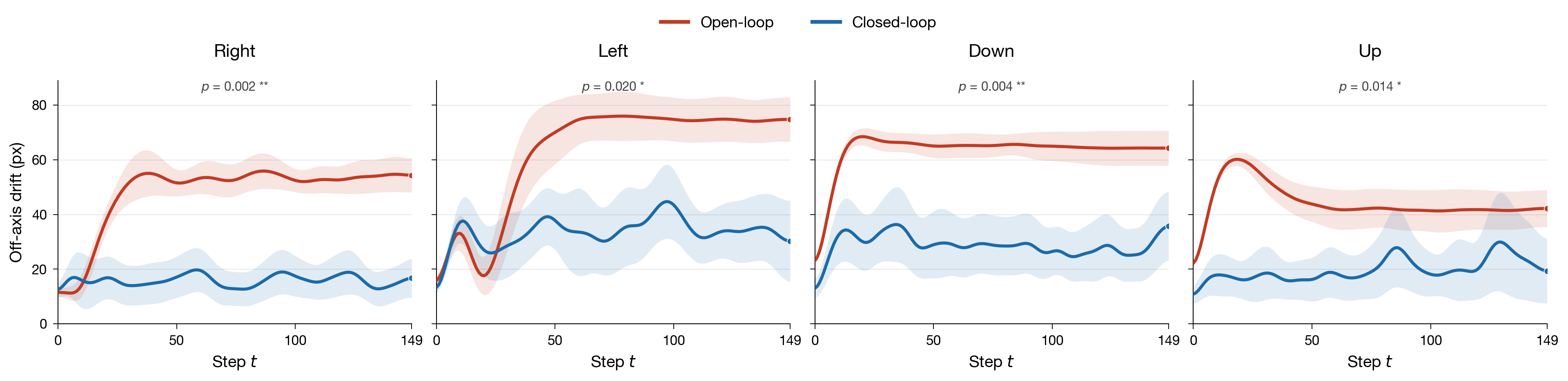}
  \caption{Visualization of the analysis.
  \textbf{Top:} Per-direction action predictions across five timesteps of a single rollout. TPM traces collapse into nearly identical paths (top row), whereas our traces fan out into four distinct directions (bottom row).
  \textbf{Bottom}: off-axis drift under open-loop vs.\ closed-loop attacker control with the same patch vocabulary. For each command, off-axis drift is the agent's vertical displacement from start under ``right''/``left'', and horizontal displacement under ``up''/``down''. Each curve is the mean over 10 paired DDIM-seed trials; the shaded band is 95\% confidence interval.}
  \label{fig:tpm_failure}
\end{figure}

We first verify that the supposed target oracle, obtained by querying the frozen victim policy model in a goal-shifted environment, cannot itself perform the new task.
Running the frozen policy without any patch in environments where the green T target is moved to four directions, yields $0/40$ task success across all four directions. 
In fact, the four actions of the victim policy given various targets were nearly identical, indicating that the policy is goal-locked. It was trained on demonstrations to a single fixed goal and generalizes poorly to novel goals.
See \Cref{fig:tpm_failure} top for visualization of the actions. The TPM actions given different targets remain identical, while our patch vocabulary consistently generates distinct action sequences.

Even granting TPM a perfect direction-specific patch, the open-loop deployment of a single fixed reference $\bar{\ba}^{\,c}$ accumulates drift along the horizon. We isolate this factor by replacing the reference oracle with our optimized patches and varying \emph{only} the deployment loop: \emph{Open-loop} holds the same direction patch on every step; \emph{Closed-loop} lets a human operator toggle and switch patches in real time. \Cref{fig:tpm_failure} bottom plots off-axis drift over 150 steps (10 paired DDIM-seed trials per direction). Open-loop drift rapidly increases, while closed-loop drift stays low in every direction with statistically significant gaps (per-direction paired Wilcoxon $p < 0.02$; combined $p{=}1.4{\times}10^{-9}$). Even with a stable action bias, open-loop alone fails to remain on-axis, while closed-loop attacker control is the structural fix.



TPM's failure is not a hyperparameter issue or a weakness of the optimization; it is a fundamental incompatibility between target-policy matching and achieving the attacker's target in a robotic context, where the latter needs a generalizable closed-loop deployment. By contrast, our objective for the universal patches does not require the victim policy to generate any out-of-distribution inputs; rather, it yields robust action-primitive biases. Moreover, the human operator plays a key role in achieving the closed-loop goal under the TAKO formulation.

\subsection{Ablation Studies}
\label{sec:ablation}
We ablate the design choices on PushT by varying patch size, perturbation budget $\varepsilon$, the number of denoising steps used during patch optimization, patch placement, agent-position conditioning, and the expectation-over-transformation (EOT) number for the diffusion noise seed. See \Cref{app:ablation} for the detailed results of all factors along with an online human verification study.

We summarize the findings as follows. (1) Small patches can still hijack direction. A $32\times32$ patch (11\% image area) reaches 94\% directional success; even $16\times16$ (2.8\%) retains 79\%. Due to the need for accurate action, PushT requires online human task verification, which reports difficulty when the patch size is below $48\times48$.
(2) Perturbation budget saturates above $\varepsilon = 64/255$. The directional shift grows monotonically with $\varepsilon$ but flattens above $64/255$ ($\delta_{\text{dir}}$ rises from 146.1 px at $64/255$ to only 187.1 px at $255/255$); $\varepsilon = 64/255$ remains a balanced default for visibility versus effectiveness. (3) The denoising chain does not erase adversarial bias. Increasing DDIM steps from 3 to 20 leaves $\delta_{\text{dir}}$ essentially flat (140.4 $\to$ 151.7 px).
(4) Patch effectiveness varies over positions. Placing the patches at the center, right edge, or bottom-right yields an effective directional bias, with a Directional success rate over $93\%$, while placing them at the top-edge and Top-left yields a weaker bias.
(5) Agent-position conditioning and EOT have a negligible effect.
\section{Conclusion}
\label{sec:conclusion}




Prior adversarial attacks on robot policies primarily ask whether an adversary can make a robot fail. This paper identifies a more serious threat: \emph{Test-time Adversarial Takeover} (TAKO), in which a human operator obtains a real-time steering interface over a victim diffusion-based policy by switching among pre-optimized universal patch vocabulary injected into the camera stream. We demonstrate this takeover interface across multiple evaluated settings, including 2D manipulation, aerial delivery, and ground navigation, in both the simulation and the physical world.

Conceptually, TAKO reframes adversarial control as an action vocabulary rather than a single target-matching perturbation. Instead of requiring the attacker to specify and track a full counterfactual target policy, a small set of reusable action primitives can be composed online to produce attacker-directed behavior. 
More broadly, these findings suggest that adversarial bias introduced through visual conditioning can persist during generative inference sufficiently to support sustained steering.
We hope this work motivates future research on stronger evaluation methods, defenses, and security mechanisms for generative robot policies under interactive adversarial control.

\textbf{Limitation} Our experiments assume white-box access for offline patch optimization and digital injection into the camera stream, and are evaluated in controlled simulation and physical setups. Realization of the takeover without these assumptions, such as via physical patches or black-box optimization, is an important direction for future work.

\bibliographystyle{unsrtnat}
\bibliography{references}

\clearpage
\newpage
\appendix
\setcounter{algorithm}{0}
\setcounter{table}{0}
\setcounter{figure}{0}

\renewcommand{\thesection}{A\arabic{section}}
\renewcommand{\thealgorithm}{A\arabic{algorithm}}
\renewcommand{\thetable}{A\arabic{table}}
\renewcommand{\thefigure}{A\arabic{figure}}
\renewcommand{\theHtable}{A\arabic{table}}
\renewcommand{\theHfigure}{A\arabic{figure}}
\onecolumn

This appendix is organized into four sections. \Cref{sec:app_setup} supplements the experimental setup of \Cref{sec:setup} with policy implementation details, patch optimization specifics, and per-task configurations. \Cref{app:ablation} provides the full data behind the ablation findings of \Cref{sec:ablation}. \Cref{app:additional} expands the patch directional success rates of \Cref{sec:patch_effectiveness} to per-direction granularity. \Cref{app:repro} reports compute resources, dataset licenses, and reproducibility information.

\section{Additional Experimental Details}
\label{sec:app_setup}

\subsection{Additional Experimental Setup}
\label{sec:add_exp_setup}

\textbf{Policy implementation details.}
All four tasks share the same underlying recipe of \citet{chi2023diffusion}---a conditional diffusion action head conditioned on visual features---but they differ in their visual encoder. PushT, SimNav, and RealNav use a ResNet-18 with GroupNorm followed by a conditional 1D UNet action head. DeliverDrone uses the NoMaD architecture~\citep{shah2023nomad}, whose visual encoder is an EfficientNet-B0 followed by a 4-layer Transformer~\citep{shah2023vint}. We load the publicly released NoMaD pretrained checkpoint and finetune on our task-specific data. We do not use NoMaD's goal-image mode, so random goal masking is also disabled during training; the policy input is image and state only, identical to vanilla Diffusion Policy. For \emph{every} task, we train policy checkpoints under all three generative paradigms---DDPM, DDIM, and flow matching~\citep{lipman2023flow}.

\textbf{Patch vocabulary optimization.}
For each task, we optimized four patches $\{\bdelta_c^*\}_{c \in \mathcal{C}}$ corresponding to the four directional commands in $\mathcal{C}$. Each patch is a square of side length equal to half the input image width, so that the patch covers 25\% of the image area in every task. The perturbation magnitude is bounded by $\varepsilon = 64/255$ in $L_\infty$ norm, which produces visible but not overwhelming artifacts. All patches are optimized using Adam~\citep{kingma2014adam} with learning rate 0.01, $\beta_1 = 0.9$, $\beta_2 = 0.999$, for 500 gradient steps with batch size 8. The observation pool $\mathcal{D}$ consists of 1000 randomly sampled frames from the training demonstrations. Task-specific deviations are noted in the per-task descriptions below. Optimization takes approximately 10 minutes per universal patch on a single NVIDIA A40 GPU. 

\subsection{Task implementation Details}
\label{sec:app_task}
\textbf{PushT} (2D manipulation): A circular agent operates in a 512$\times$512 pixel workspace and must push a T-shaped block to a fixed target pose. The policy observes two consecutive 96$\times$96 RGB frames and the agent's 2D position, and predicts 16 steps of 2D pixel-space velocity, of which 8 are executed per planning cycle. This task serves as the primary concept validation environment.

\paragraph{PushT policy.}
We train the Diffusion Policy of \citet{chi2023diffusion} for 300 epochs on the standard PushT demonstration dataset (200 episodes). The vision encoder is ResNet-18 with GroupNorm (replacing BatchNorm for small-batch inference stability). The denoiser is a conditional 1D UNet with 256 base channels and attention at 8$\times$ downsampling. DDPM training uses 100 diffusion steps; DDIM inference uses 10 steps by default. The observation horizon is 2 frames; the prediction horizon is 16 steps; the action execution horizon is 8 steps.

\textbf{DeliverDrone} (simulated aerial delivery): A simulated quadrotor navigates a ${\sim}$10$\times$10\,m outdoor terrain in Gazebo Harmonic to deliver a package to a blue landing pad at (3.0, 0.0)\,m. The drone observes four stacked 96$\times$96 top-down RGB frames, a 3-dimensional ego-velocity state, and a pre-captured goal image of the landing area. The policy outputs 8 XY waypoints in metric coordinates. This task instantiates a realistic intelligent logistics scenario where camera-stream compromise could enable adversarial rerouting of autonomous deliveries.

\paragraph{DeliverDrone policy.}
We follow the NoMaD~\citep{shah2023nomad} codebase. The vision encoder is EfficientNet-B0 pretrained on ImageNet, followed by a 4-layer Transformer with 256-dimensional embeddings and 4 attention heads. The diffusion action head uses a 3-layer MLP denoiser with 256 hidden units. We initialize from the publicly released NoMaD checkpoint and finetune for 50 epochs on 100 task-specific demonstration trajectories collected by the authors; we do not redo the original NoMaD pretraining. The observation horizon is 4 frames; the prediction horizon is 8 waypoints.

\textbf{SimNav} (simulated ground navigation): An FPS agent navigates a bedroom/living-room environment in AI2-THOR~\citep{kolve2017ai2thor} (Unity 2021.3) to reach a goal position specified by XZ coordinates. The policy observes two consecutive 256$\times$256 first-person-view (FPV) RGB frames and the agent's 2D position, and predicts 16 steps of continuous velocity $(v, \omega)$ (linear and angular), of which 8 are executed per planning cycle. The environment communicates via a dm\_env\_rpc~\citep{dm_env_rpc2019,ward2020using} gRPC bridge on the Unity side, enabling real-time closed-loop control. This task tests the attack under egocentric visual observations with high-resolution input and continuous velocity control---a distinct perceptual and actuation regime from the allocentric top-down views of PushT and DeliverDrone.

\paragraph{SimNav policy.}
SimNav uses the same architecture as PushT---ResNet-18 vision encoder (GroupNorm) with a conditional 1D UNet---but trained via flow matching rather than DDPM. The policy maps 2$\times$256$\times$256 FPV RGB frames and a 2D agent position to 16 steps of continuous velocity $(v, \omega)$, executing 8 per cycle. Training data consists of human mouse demonstrations collected in AI2-THOR via a custom dm\_env\_rpc bridge. We merge the original forward-goal collection with a reverse-goal collection, yielding a zarr replay buffer with 152 episodes and 13{,}518 frames; the merged policy is trained for 100 epochs and reaches final training loss 0.13. Policy evaluation uses an 8-step forward-Euler ODE solve. Patch optimization uses lr$=$1.0 (raw $[0,255]$ pixel input), 128$\times$128 center patch ($25\%$ of 256$\times$256 image), raw-pixel $\varepsilon=64$, 300 steps with batch size 8 drawn from a pool of 300 observations. Patches are optimized with 3-step flow inference and evaluated with 8-step inference.

\textbf{RealNav} (physical ground navigation): An iRobot Create~3 chassis navigates an indoor corridor using only a forward-facing RGB camera and onboard odometry as policy inputs. The deployed policy uses the same image-based Diffusion Policy family as PushT---a ResNet-18 vision encoder with GroupNorm and a conditional 1D UNet action head---but the final online model is trained with flow matching. The policy observes two consecutive 96$\times$96 FPV RGB frames and the robot pose, and predicts 16 steps of 2D body-frame waypoints, of which 8 are executed per planning cycle. This task validates the attack on a physical platform where sensor latency, lighting variation, wireless ROS~2 communication, and real-world robot dynamics are present.

\paragraph{RealNav policy.}
The RealNav policy is trained from 30 human-teleoperated corridor trajectories collected on the physical robot. The raw bags contain 23{,}376 camera images and 764\,s of driving, covering 97.8\,m of total path length. We resample the data to 4\,Hz and obtain 3{,}031 processed frames, split into 27 training trajectories and 3 held-out test trajectories. Input images are stored as 96$\times$96 RGB frames; during training the encoder applies a 76$\times$76 random crop. The action dimension is 2, representing a body-frame waypoint $(\Delta x,\Delta y)$. The observation horizon is 2 frames, the prediction horizon is 16 waypoints, and the execution horizon is 8 waypoints.

The final online RealNav policy is a flow-matching variant of Diffusion Policy with a ResNet-18 GroupNorm visual encoder and a conditional 1D UNet action model. It is trained for 400 epochs with AdamW (lr $=10^{-4}$, batch size 8, cosine schedule, 50 warmup steps). Flow-matching inference uses 4 Euler steps. In deployment, the policy runs at 8\,Hz on the workstation, selects a body-frame waypoint from the predicted sequence, and converts it to differential-drive velocity commands through a waypoint controller. The robot-side Raspberry Pi publishes camera and RPLIDAR streams; the RPLIDAR is used only by an emergency-stop node and is not consumed by the policy.

Patch optimization uses Adam with lr$=$0.01 for 500 steps on normalized $[0,1]$ RGB input, a 48$\times$48 center patch (25\% of the 96$\times$96 policy input), $\varepsilon = 64/255$, batch size 8, and an observation pool sampled from the RealNav corridor training set.



\paragraph{RealNav Physical platform.}
RealNav runs on an iRobot Create~3 chassis with a Raspberry~Pi~4B companion computer (Ubuntu~22.04 LTS, ROS~2 Humble). The Pi carries a single USB-attached 170-degree wide-angle camera that publishes 640$\times$480 YUYV frames at 30\,Hz on \texttt{/camera/image\_raw/compressed}; this is the only sensor consumed by the policy. The camera is mounted on an aluminum mast approximately 42\,cm above the robot base and 15\,cm behind the robot center, and its intrinsics are calibrated with an equidistant fisheye model. The Create~3 contributes its on-board odometry and bumper/cliff switches; an RPLIDAR is also present but is wired only to an emergency-stop node for obstacle safety and is not provided to the policy. 

\begin{figure}[t]
  \centering
  \includegraphics[width=\linewidth]{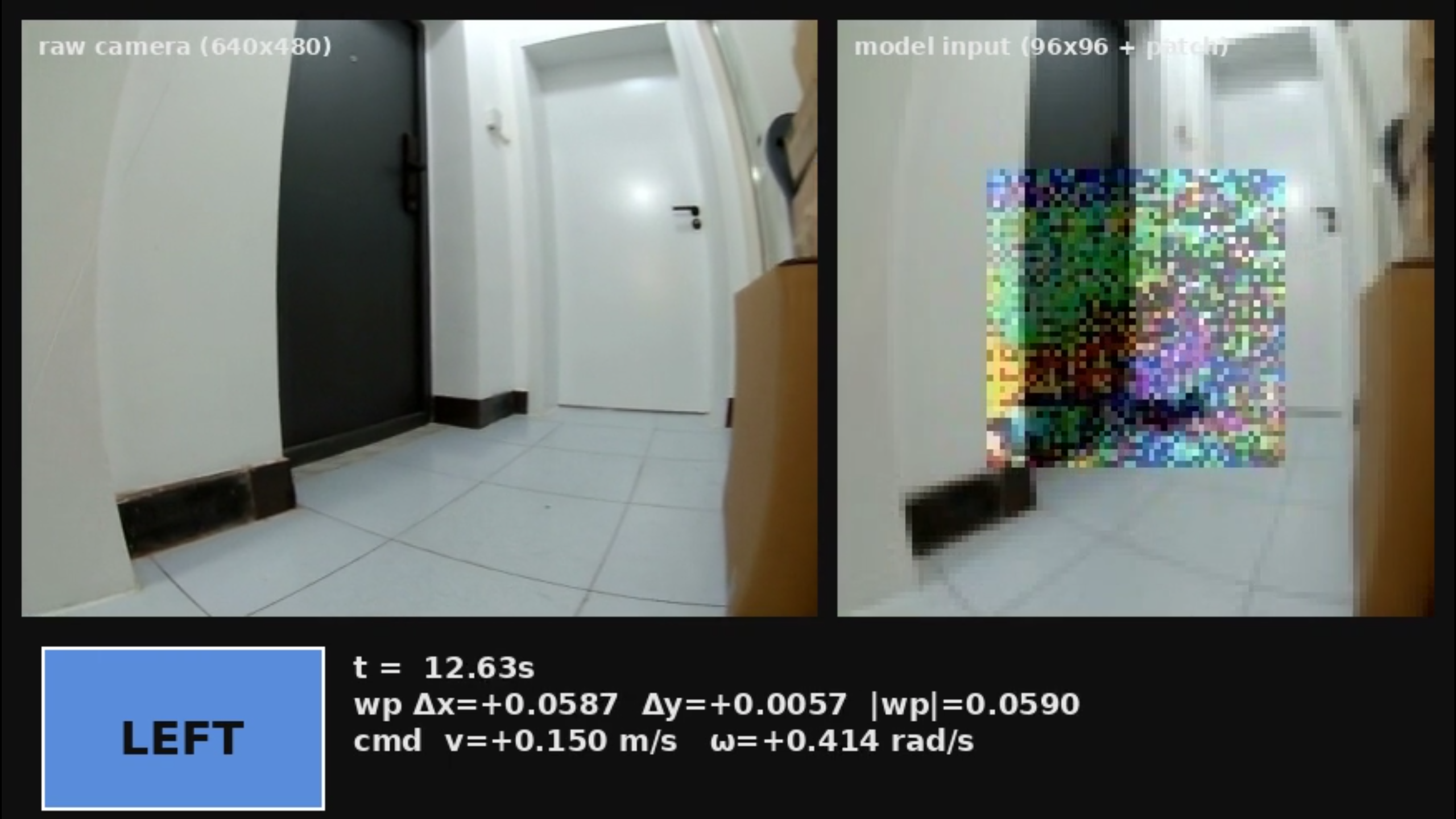}
  \setlength{\abovecaptionskip}{1pt}
  \caption{RealNav operation interface. Left: raw camera view. Right: model input with the patch. The “Left” label indicates the current attack direction, and the side panel shows the robot’s real-time status.}
  \label{fig:realnav}
\end{figure}

\begin{figure}[t]
  \centering
  \includegraphics[width=\linewidth]{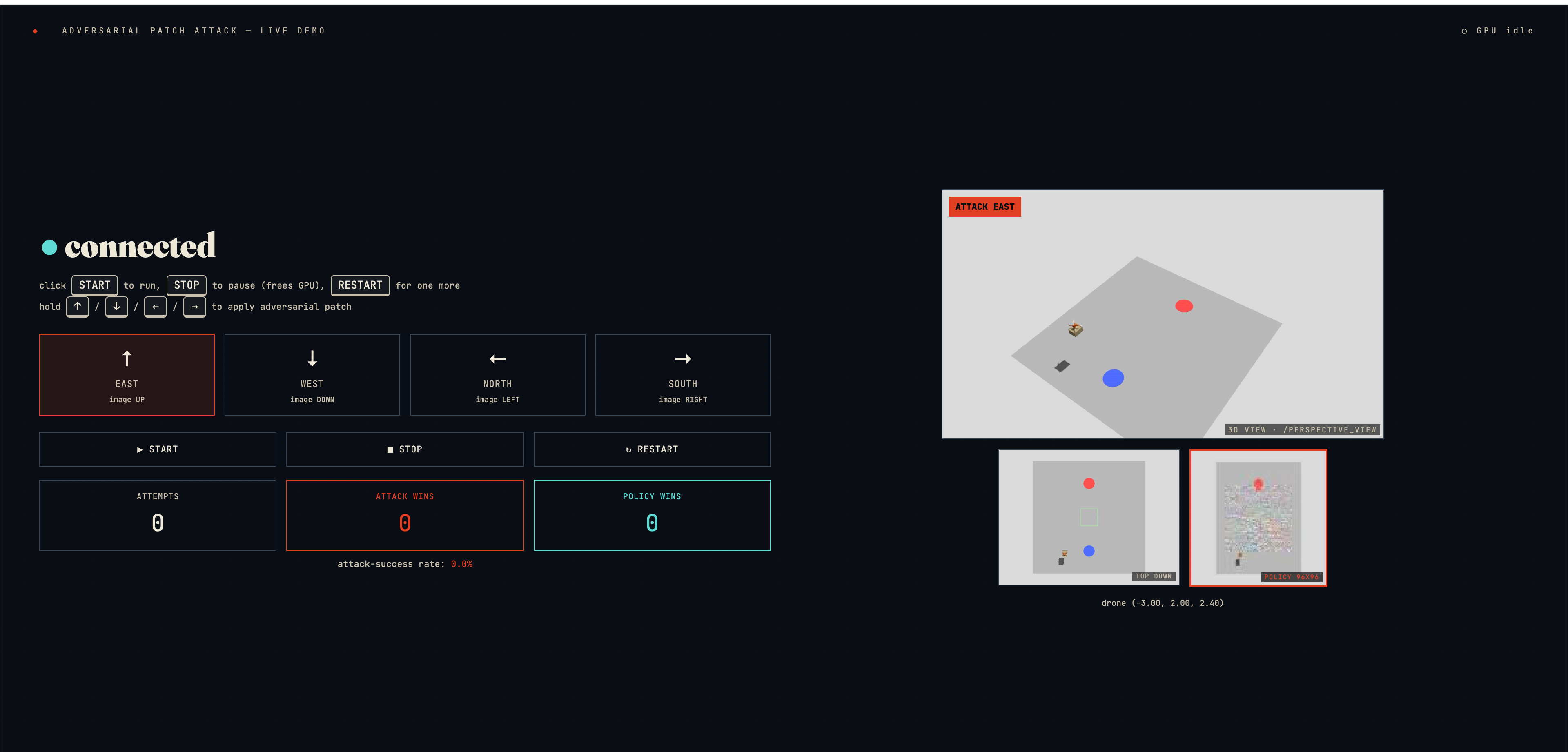}
  \setlength{\abovecaptionskip}{1pt}
  \caption{Drone operation interface.  Left: operator instructions. Right: raw camera view with or without the patch. The same interface is used for the SimNav and PushT experiments.}
  \label{fig:drone}
\end{figure}

\paragraph{Human-in-the-loop Human Operation Interface}
We built human-in-the-loop operation interfaces for all four experiments, through which an operator can control the victim policy by switching directional adversarial patches in real time. PushT uses a FastAPI + WebSocket backend that streams the simulation view and control to a web frontend directly. DeliverDrone bridges ROS\,2 topics to the web frontend via rosbridge\_server. SimNav communicates with Unity through the dm\_env\_rpc gRPC bridge, with a Python backend forwarding frames to the web frontend. In RealNav, the interface operates on the live camera stream (Figure~\ref{fig:realnav}). The web-based interfaces for the three simulated tasks are shown in Figure~\ref{fig:drone}.

\section{Additional Ablation Study}
\label{app:ablation}

\begin{table}[ht]
  \caption{Ablation summary on PushT. \emph{(a)} Patch size sweep at $\varepsilon = 64/255$, center placement: even an 11\%-area patch hijacks direction at 94\%. \emph{(b)} Perturbation budget at $48\times48$, center placement: directional shift saturates above $\varepsilon = 64/255$. \emph{(c)} DDIM denoising steps used during patch optimization, at $48\times48$ and $\varepsilon=64/255$: directional shift is essentially flat from 3 steps onward, indicating that the denoising chain does not attenuate adversarial bias in the conditioning signal. \emph{(d)} Patch placement at $32\times32$, target direction = right: center is the direction-agnostic optimum. Default configurations are shown in \textbf{bold}.}
  \label{tab:ablation_summary}
  \centering
\small
\vspace{4pt}
 
\begin{minipage}[t]{0.48\textwidth}
\centering
\textit{(a) Patch size}\par\vspace{3pt}
\begin{tabular}{@{}l S[table-format=3.0] S S@{}}
\toprule
{Patch size} & {Dir.\ rate (\%)} & {$\delta_{\mathrm{dir}}$ (px)} & {$\|\Delta\|_2$ (px)} \\
\midrule
$16 \times 16$ &  79 &  37.1 &  61.4 \\
$32 \times 32$ &  94 &  97.8 & 128.3 \\
\rowcolor{black!5}
\B{$48 \times 48$} & \B{98} & \B{146.1} & \B{172.5} \\
$64 \times 64$ &  99 & 213.7 & 241.2 \\
$96 \times 96$ & 100 & 294.5 & 312.8 \\
\bottomrule
\end{tabular}
\end{minipage}
\hfill
\begin{minipage}[t]{0.48\textwidth}
\centering
\textit{(b) Perturbation budget $\varepsilon$}\par\vspace{3pt}
\begin{tabular}{@{}l S[table-format=3.0] S S@{}}
\toprule
{$\varepsilon$} & {Dir.\ rate (\%)} & {$\delta_{\mathrm{dir}}$ (px)} & {$\|\Delta\|_2$ (px)} \\
\midrule
$16/255$  &  71 &  13.4 &  28.9 \\
$32/255$  &  89 &  62.3 &  85.7 \\
\rowcolor{black!5}
\B{$64/255$}  & \B{98} & \B{146.1} & \B{172.5} \\
$128/255$ &  99 & 178.4 & 203.1 \\
$255/255$ & 100 & 187.1 & 211.6 \\
\bottomrule
\end{tabular}
\end{minipage}
 
\vspace{14pt}
 
\begin{minipage}[t]{0.48\textwidth}
\centering
\textit{(c) Denoising steps}\par\vspace{3pt}
\begin{tabular}{@{}l S[table-format=3.0] S S@{}}
\toprule
{DDIM steps} & {Dir.\ rate (\%)} & {$\delta_{\mathrm{dir}}$ (px)} & {$\|\Delta\|_2$ (px)} \\
\midrule
1  &  78 &  52.3 &  89.1 \\
3  &  97 & 140.4 & 168.2 \\
5  &  98 & 143.8 & 170.9 \\
\rowcolor{black!5}
\B{10} & \B{98} & \B{146.1} & \B{172.5} \\
20 &  98 & 151.7 & 178.3 \\
\bottomrule
\end{tabular}
\end{minipage}
\hfill
\begin{minipage}[t]{0.48\textwidth}
\centering
\textit{(d) Patch placement at $32 \times 32$}\par\vspace{3pt}
\begin{tabular}{@{}l S[table-format=3.0] S S@{}}
\toprule
{Placement} & {Dir.\ rate (\%)} & {$\delta_{\mathrm{dir}}$ (px)} & {$\|\Delta\|_2$ (px)} \\
\midrule
\rowcolor{black!5}
\B{Center}   & \B{93}  & \B{96.6}  & \B{117.4} \\
Right-edge   & 100 & 119.7 & 123.7 \\
Bottom-right & 100 &  87.7 & 107.0 \\
Top-edge     &  77 &  29.1 &  57.2 \\
Top-left     &  89 &   6.4 &   8.4 \\
\bottomrule
\end{tabular}
\end{minipage}
\end{table}

This section provides the full data behind the ablation findings of \Cref{sec:ablation}. \Cref{tab:ablation_summary} reports the four main sensitivity factors (patch size, perturbation budget, denoising steps, patch placement). \Cref{tab:ablation_position} and \Cref{tab:ablation_eot} report the agent-position conditioning and EOT ablations, respectively. \Cref{tab:online_verification} pairs offline metrics with online human operation outcomes to verify that offline metrics predict online controllability. All experiments are on PushT, varying one factor at a time with defaults held at patch size $48\times48$, $\varepsilon = 64/255$, DDIM 10 steps, evaluated on 100 held-out observations unless otherwise stated. We report three metrics: $\delta_{\mathrm{dir}}$ (mean action shift along the target direction, in pixels), directional success rate (fraction of samples shifted in the correct direction), and $\|\Delta\|_2$ (total $L_2$ action displacement).

\paragraph{Agent-position conditioning.}
The PushT policy reads a 2D agent position alongside the visual observation. We test whether patch effectiveness relies on a specific position input by replacing the true position with a fixed value (center / corner) or with a uniformly random value drawn at each call. \Cref{tab:ablation_position} reports a $\le 6\%$ change in $\delta_{\mathrm{dir}}$ across all three settings, confirming that the visual perturbation dominates over the proprioceptive anchor: the attack does not need any specific position input to succeed.

\begin{table}[h]
  \caption{Agent-position conditioning ablation on PushT (fixed patch size $48\times48$, $\varepsilon = 64/255$, DDIM 10 steps).}
  \label{tab:ablation_position}
  \centering
  \begin{tabular}{lccc}
    \toprule
    Position conditioning & $\delta_{\mathrm{dir}}$ (px) & {Dir.\ rate (\%)} & $\|\Delta\|_2$ (px) \\
    \midrule
    Fixed (center)   & 146.1 & 98 & 172.5 \\
    Fixed (corner)   & 145.4 & 97 & 170.8 \\
    Random (uniform) & 154.3 & 98 & 181.2 \\
    \bottomrule
  \end{tabular}
\end{table}

\paragraph{Expectation-over-transformation (EOT).} 
We increase the EOT seed batch from 1 to 8 during patch optimization. \Cref{tab:ablation_eot} shows no measurable improvement, consistent with the near-deterministic DDIM sampling ($\eta=0$) used throughout.

\begin{table}[h]
  \caption{EOT ablation on PushT (fixed patch size $48\times48$, $\varepsilon = 64/255$, DDIM 10 steps).}
  \label{tab:ablation_eot}
  \centering
  \begin{tabular}{lccc}
    \toprule
    EOT seeds & $\delta_{\mathrm{dir}}$ (px) & {Dir.\ rate (\%)} & $\|\Delta\|_2$ (px) \\
    \midrule
    1 (fixed) & 141.0 & 98 & 167.3 \\
    4         & 140.2 & 97 & 166.8 \\
    8         & 139.4 & 98 & 165.9 \\
    \bottomrule
  \end{tabular}
\end{table}

\paragraph{Online human-in-the-loop verification.} The takeover claim is fundamentally about whether a human operator can drive the policy in real time. We therefore validate that the offline metrics ($\delta_{\mathrm{dir}}$ and directional success rate) predict actual online controllability. We train patches at four sizes ($16\times16$ to $48\times48$, $\varepsilon = 64/255$, DDIM 3 steps) and test them in the interactive PushT demo, where the operator uses arrow keys to switch attack direction in real time. \Cref{tab:online_verification} reports two binary online outcomes: \emph{direction hijack} (operator can steer the agent in the commanded direction) and \emph{task completion} (operator can push the T-block to a chosen target region).

\begin{table}[h]
  \caption{Online human-in-the-loop verification of offline metrics on PushT. The composite $\text{dir.\ rate} \times \delta_{\mathrm{dir}}$  predicts online controllability better than either factor alone.}
  \label{tab:online_verification}
  \centering
  \begin{tabular}{lcccccc}
    \toprule
    Patch size & Area & $\delta_{\mathrm{dir}}$ & {Dir.\ rate (\%)} & Composite & Dir.\ hijack & Task compl. \\
    \midrule
    $16\times16$ & 2.8\%  & 37  & 79 & 29  & \ding{55} & --- \\
    $32\times32$ & 11.1\% & 93  & 94 & 87  & \ding{51} & Difficult \\
    $48\times48$ & 25.0\% & 146 & 98 & 143 & \ding{51} & \ding{51} \\
    \bottomrule
  \end{tabular}
\end{table}

Two thresholds emerge. \emph{Direction hijack} requires $\text{dir.\ rate} \times \delta_{\mathrm{dir}} \gtrsim 50\text{--}87$, met by $32\times32$ patches (11\% image area). \emph{Task completion} requires $\text{dir.\ rate} \times \delta_{\mathrm{dir}} \gtrsim 100$, met by $48\times48$ patches (25\% area). Directional success rate alone is insufficient: a $16\times16$ patch reaches 79\% directional success rate but its $\delta_{\mathrm{dir}} = 37$\,px is too small relative to the 512-px action space to be operator-controllable. This validates the choice of $48\times48$ as the main-paper default and connects the offline ablations of \Cref{sec:ablation} to the takeover claim of \Cref{sec:main_results}.

\section{Additional Experimental Results}
\label{app:additional}

This section expands the headline numbers of \Cref{sec:patch_effectiveness} to per-direction granularity, supporting the cross-task transferability claim under each task's native action representation.

\paragraph{Per-direction transferability across tasks and inference families.}
\Cref{tab:inference_ablation} expands \Cref{tab:patch} to the per-direction granularity. Each value is the fraction of evaluation samples whose predicted action chunk shifts along the commanded direction under that task's native action representation. The relevant comparison is row-to-row within each task block; absolute success rates are not directly comparable across tasks because the action spaces differ (Cartesian/waypoint versus velocity $(v, \omega)$). Across every task and direction the optimized vocabulary preserves directional controllability under all three inference families, supporting the main-paper claim that the vulnerability resides in the visual conditioning pathway rather than in a specific denoiser.

\begin{table}[h]
  \caption{Per-direction directional success rate (\%) across tasks and inference families. SimNav uses velocity-axis success over $(v,\omega)$; other tasks use Cartesian/waypoint action axes. Comparisons are within each task block.}
  \label{tab:inference_ablation}
  \centering
  \small
  \begin{tabular}{lcccc}
    \toprule
    Task / Inference & Right & Left & Forward & Backward \\
    \midrule
    \multicolumn{5}{l}{\emph{Cartesian/waypoint action outputs}} \\
    \multicolumn{5}{l}{\textbf{PushT}} \\
    \;\;DDIM (3 steps) & 97.7 & 100 & 98.4 & 98.4 \\
    \;\;DDPM (100 steps) & 97.7 & 100 & 99.2 & 97.7 \\
    \;\;Flow matching (Euler-10) & 98.4 & 100 & 99.2 & 97.7 \\
    \midrule
    \multicolumn{5}{l}{\textbf{DeliverDrone}} \\
    \;\;DDIM (10 steps) & 100 & 100 & 100 & 100 \\
    \;\;DDPM (10 steps) & 100 & 98.0 & 100 & 100 \\
    \;\;Flow matching & 100 & 100 & 100 & 100 \\
    \midrule
    \multicolumn{5}{l}{\textbf{RealNav}} \\
    \;\;Flow matching (Euler-4) & 99.0 & 92.0 & 98.0 & 100 \\
    \;\;DDIM (3 steps) & 71.0 & 79.0 & 67.0 & 81.0 \\
    \;\;DDPM (100 steps) & 77.0 & 72.0 & 66.0 & 87.0 \\
    \midrule
    \multicolumn{5}{l}{\emph{Velocity-control action output}} \\
    \multicolumn{5}{l}{\textbf{SimNav} (Right$=$turn\_right, Left$=$turn\_left)} \\
    \;\;Flow matching (8-step eval) & 83.6 & 75.8 & 70.3 & 74.2 \\
    \;\;DDIM (3-step patches) & 85.7 & 69.3 & 63.3 & 83.0 \\
    \bottomrule
  \end{tabular}
  \\[2pt]
  {\footnotesize Values are directional success rates (\%) under each task's native action parameterization.}
\end{table}

\section{Reproducibility, Resources, and Datasets}
\label{app:repro}

This section covers compute, datasets and their licenses, code release, and the broader-impact disclosure for the experiments and the attack pipeline.

\paragraph{Compute resources.}
The pipeline has two distinct compute axes, both modest. \emph{Policy training and patch optimization} are run on a server-side NVIDIA A40 GPU. Policy training reuses the recipes of \citet{chi2023diffusion} (PushT, SimNav, RealNav) and \citet{shah2023nomad} (DeliverDrone) and scales with dataset size; no single policy in our experiments exceeds 4 GPU-hours of training. Patch optimization produces one $48\times48$ universal directional patch in roughly 10 minutes (500 Adam steps, batch size 8, DDIM 10 unrolled denoising steps), so the full four-direction vocabulary per task takes well under one GPU-hour. The attack does not update policy weights, so policy-training cost is amortized across all reported takeover and TPM-baseline runs. \emph{Real-time inference for the RealNav physical-robot deployment} is run on a separate mobile graphics workstation with an NVIDIA RTX 4090 GPU, which the policy uses at 8\,Hz over the wireless link to the Raspberry Pi described in \Cref{sec:app_setup}. Aggregating across all four tasks and three inference families, the reported experiments consume roughly 60 A40-hours plus about 10 RTX~4090-hours; including preliminary and discarded runs, total project compute is approximately three times this figure.

\paragraph{Datasets and licenses.}
We list each external asset together with its license and source.
PushT demonstrations are the public 200-episode dataset released with the Diffusion Policy code base~\citep{chi2023diffusion} (MIT License).
SimNav uses AI2-THOR scenes~\citep{kolve2017ai2thor} (Apache 2.0); training trajectories are collected by the authors via mouse teleoperation in the \texttt{FloorPlan\_Train1\_1} bedroom/living-room scene.
DeliverDrone uses Gazebo Harmonic (Apache 2.0) and the publicly released NoMaD pretrained checkpoint~\citep{shah2023nomad} (MIT License); the finetuning data are author-collected demonstration trajectories. We use only the released checkpoint and do not redistribute or retrain on its underlying corpus, so any restrictions on those datasets do not propagate to our use.
RealNav corridor demonstrations are collected by the authors with an iRobot Create~3 in an indoor corridor; no human subjects beyond the authors are involved. The robot stack uses the iRobot Create~3 ROS~2 SDK (BSD-3-Clause) and ROS~2 Humble (Apache 2.0).
The training and patch-optimization code is built on PyTorch (BSD-3-Clause) and the Diffusion Policy reference implementation~\citep{chi2023diffusion} (MIT License). All cited assets are used in accordance with the terms of their respective licenses.

\paragraph{Code, patches, and reproducibility.}
We will release the patch optimization code, the four directional patches per task, the evaluation observation pools backing \Cref{tab:main_results} and \Cref{tab:inference_ablation}, and the attack/evaluation scripts on a public GitHub repository upon acceptance. The release is intentionally deferred past the review window to preserve author anonymity; we therefore do not bundle the code into the supplementary material. The takeover-trial protocol (10 trials per operator per task, 4 operators, blue-marker target) is fully specified in \Cref{sec:setup}; online human trials are inherently non-deterministic, but offline patch optimization and offline evaluation are seeded for reproducibility.

\paragraph{Broader impacts and responsible disclosure.}
This paper documents an attack that converts deployed diffusion-based robot policies into operator-steerable instruments via test-time visual perturbation. The threat is structural to the visual conditioning pathway and is not specific to any single deployed system. We disclose it now---rather than withholding it---because awareness is a prerequisite for defenses, and because the necessary attack components (universal adversarial patches, differentiable diffusion inference, white-box policy access) are individually present in the public literature and can be assembled by any sufficiently motivated party. We deliberately limit our experiments to controlled simulation and authors' own physical laboratory; we do not attack any third-party deployed robotic system. Plausible defensive directions include input-space anomaly detection on the camera stream, certified robustness on the visual encoder, and policy-architecture changes that diversify or randomize the visual conditioning pathway during inference. We hope the framing of takeover-vs-disruption developed here helps future work measure robotic policy robustness against the right adversary model.

\end{document}